\documentclass{article}
                                  \usepackage[utf8]{inputenc}
\usepackage[T1]{fontenc}
\usepackage{microtype}
\usepackage[table,svgnames]{xcolor}
\usepackage{graphicx}
\usepackage{amsmath}
\usepackage{amssymb}
\usepackage{multirow}
\usepackage{mathtools}

\usepackage[noend]{algorithmic}
\PassOptionsToPackage{hyphens}{url}\usepackage[hidelinks]{hyperref}
\usepackage[accepted]{icml2020}
\usepackage{bm}
\usepackage{bbm}
\usepackage{booktabs}
\usepackage{cleveref}
\usepackage{subcaption}
\usepackage{adjustbox}
\renewcommand{\vec}[1]{\bm{#1}}

\newcommand{\Set}[1]{\mathcal{#1}}
\newcommand{\hist}{\Set{H}}
\newcommand{\LL}{\mathcal{L}}
\newcommand{\R}{\mathbb{R}}
\newcommand{\vu}{\vec{u}}
\newcommand{\vv}{\vec{v}}
\newcommand{\param}{\theta}
\newcommand{\vpar}{\vec{\param}}

\newcommand{\vs}[1]{\smash{\widehat{\vec{z}}}\,\vphantom{z}^{#1}}

\icmltitlerunning{Learning Multivariate Hawkes Processes at Scale}
\date{}
\title{Learning Multivariate Hawkes Processes at Scale}
\hypersetup{
 pdfauthor={Maximilian Nickel},
 pdftitle={Learning Multivariate Hawkes Processes at Scale},
 pdfkeywords={},
 pdfsubject={},
 pdfcreator={Emacs 26.3 (Org mode 9.4)}, 
 pdflang={English}}
\begin{document}

\twocolumn[
\icmltitle{Learning Multivariate Hawkes Processes at Scale}

% It is OKAY to include author information, even for blind
% submissions: the style file will automatically remove it for you
% unless you've provided the [accepted] option to the icml2019
% package.

% List of affiliations: The first argument should be a (short)
% identifier you will use later to specify author affiliations
% Academic affiliations should list Department, University, City, Region, Country
% Industry affiliations should list Company, City, Region, Country

% You can specify symbols, otherwise they are numbered in order.
% Ideally, you should not use this facility. Affiliations will be numbered
% in order of appearance and this is the preferred way.
\icmlsetsymbol{equal}{*}

\begin{icmlauthorlist}
\icmlauthor{Maximilian Nickel}{fair}
\icmlauthor{Matthew Le}{fair}
\end{icmlauthorlist}

\icmlaffiliation{fair}{Facebook AI Research, New York, New York, USA}

\icmlcorrespondingauthor{Maximilian Nickel}{\texttt{maxn@fb.com}}:

% You may provide any keywords that you
% find helpful for describing your paper; these are used to populate
% the "keywords" metadata in the PDF but will not be shown in the document
\icmlkeywords{Machine Learning, ICML}

\vskip 0.3in
]

\printAffiliationsAndNotice{}  % leave blank if no need to mention equal contribution

\begin{abstract}
Multivariate Hawkes Processes (MHPs) are an important class of temporal point
processes that have enabled key advances in understanding and predicting social
information systems. However, due to their complex modeling of temporal
dependencies, MHPs have proven to be notoriously difficult to scale, what has
limited their applications to relatively small domains. In this work, we propose
a novel model and computational approach to overcome this important limitation.
By exploiting a characteristic sparsity pattern in real-world diffusion
processes, we show that our approach allows to compute the \emph{exact} likelihood
and gradients of an MHP -- independently of the ambient dimensions of the
underlying network. We show on synthetic and real-world datasets that our model
does not only achieve state-of-the-art predictive results, but also improves
runtime performance by multiple orders of magnitude compared to standard methods
on sparse event sequences. In combination with easily interpretable latent variables
and influence structures, this allows us to analyze diffusion processes at
previously unattainable scale.
\end{abstract}

\section{Introduction}
\label{sec:intro}
Time, and in particular the magnitude of time intervals between events, carries
important information about latent structures of event sequences. Temporal
point processes (TPPs) are a flexible and powerful paradigm for modeling such
discrete event sequences that are localized in continuous time. Consequently,
TPPs have found important applications in diverse fields such as social and
information systems \citep{gomez-rodriguez2011uncovering,iwata2013discovering},
human mobility and learning
\citep{jankowiak2017uncovering,mavroforakis2017modeling}, finance
\citep{bacry2015hawkes}, health \cite{alaa2017learning}, and recommender systems
\citep{jing2017survival,kumar2019predicting}.

\begin{figure}[t]
\centering
\begin{equation*}
\lambda_x(t|h) = f \begin{pmatrix}
\mu & \alpha \\
\mathord{\includegraphics[height=7em]{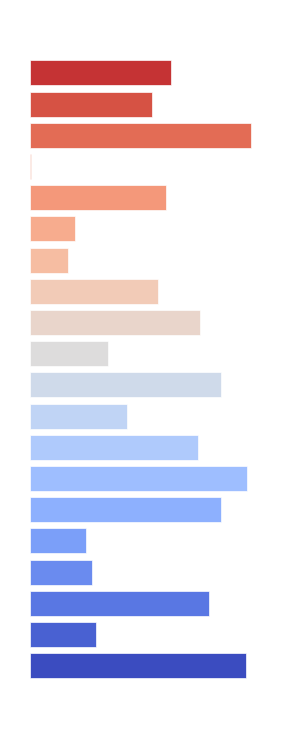}} \raisebox{1ex}{\ ,} &
\mathord{\includegraphics[height=7em]{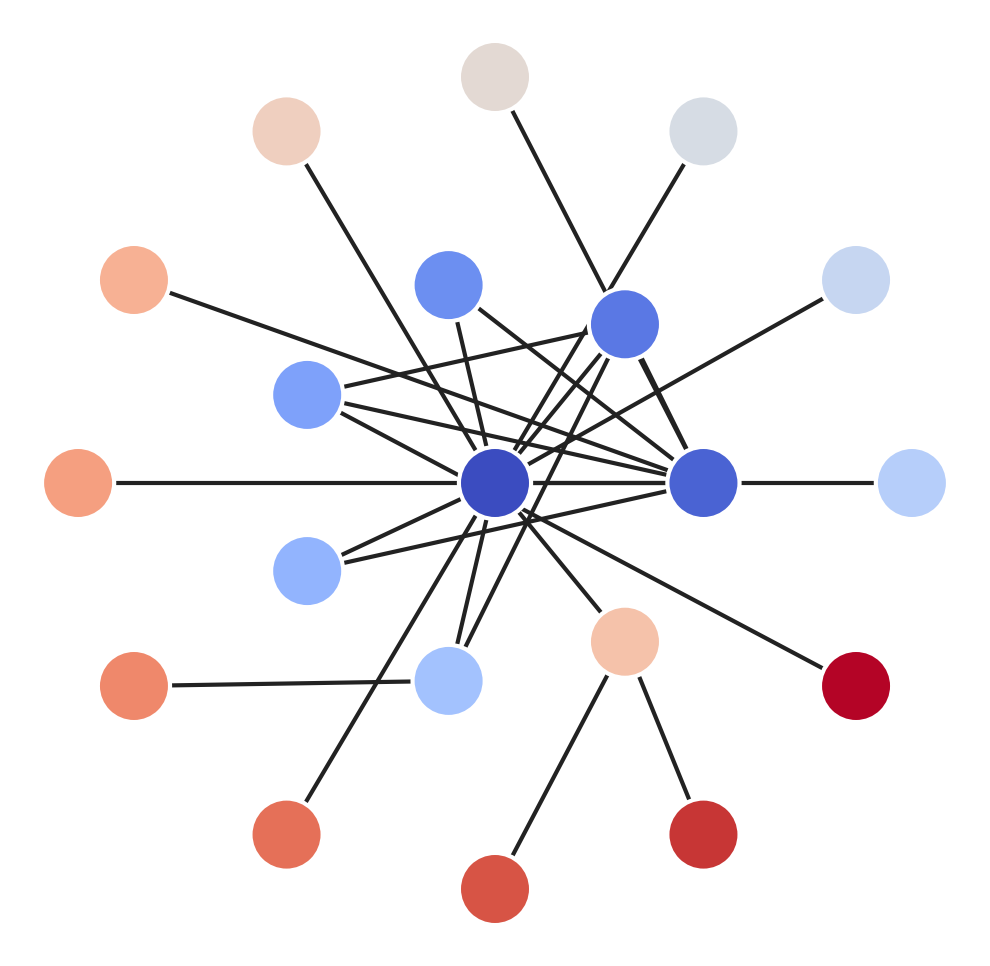}}
\end{pmatrix}
\end{equation*}

{\footnotesize \hspace{1em} Inference {\normalsize $\uparrow \downarrow$} Generation}

\includegraphics[width=.95\columnwidth]{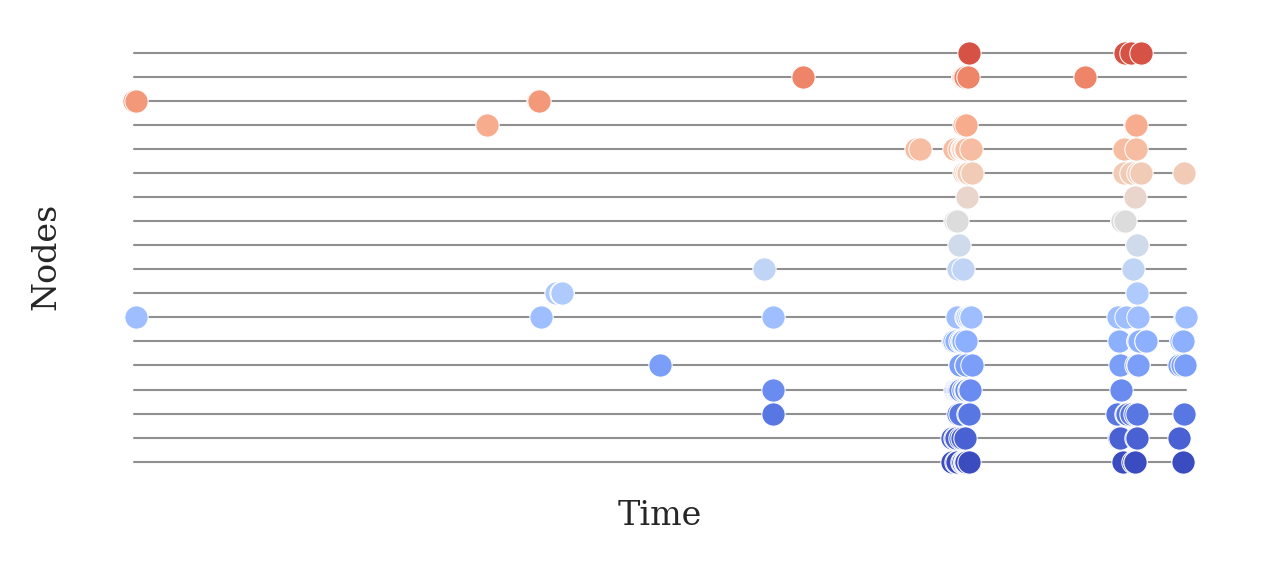}
\caption{\small Multivariate Hawkes Processes model inter-dependent events along
multiple entities in continuous time. Model parameters such as
base intensities ($\mu$) and mutual excitation ($\alpha$) can be inferred from
observed episodes and provide insights into the latent structure of event
sequences.\label{fig:mhp-example}}
\vspace{-1em}
\end{figure}

Hawkes Processes \citep{hawkes1971spectra} are an important class of TPPs which
are widely used for modeling temporal events with self-exciting behaviour, i.e.,
processes were the occurrence of an event increases the likelihood of another
event happening in the near future.
Multivariate Hawkes Processes (MHPs) extend this approach by also accounting for
the mutual excitation of events along different entities (or dimensions).
Such self- and mutually-exciting processes are central to
modeling a wide range of real-world phenomena. For instance, Hawkes Processes
have been found to be an excellent approach for modeling the self-exciting and
bursty nature of information cascades in social networks \citep{zhao2015seismic};
for modeling the mutually exciting and retaliatory patterns in gang violence
\citep{linderman2014discovering}; for modeling earthquake aftershock sequences
\citep{ogata1998space}; and have been proposed to mitigate the spread of fake news
\citep{farajtabar2017fakenews}.

Moreover, a distinctive advantage of MHPs is that their model parameters are
readily interpreable and can provide valuable insights into the latent structure
of event sequences (see also \Cref{fig:mhp-example}). This has, for instance, been
exploited to infer latent influence strucures from diffusion processes
\citep{boehmke2018new}, to connect Hawkes processes with epidemiological models
\citep{rizoiu2018sir}, and to reveal causal relations between entities from
temporal events \citep{xu2016grangermhp,achab2017uncovering}.

Despite this diverse range of applications and appealing properties, wider
adoption of MHPs has been hindered due to a fundamental limitation: existing
models and inference methods for MHPs have proven to be very difficult to scale
due to their complex modeling of temporal dependencies. For this reason, MHPs
have typically been applied only to relatively small domains, e.g., diffusion
networks with a few hundred or thousand nodes (although the number of events
can be larger). A common strategy for larger domains is often to limit the
number of entities by subsampling the dataset.  This not only reduces the
coverage of the model, but can also lead to incorrectly estimated influence
structures as subsampling affects a process' latent relational structures.
Altogether, this leads to an unfortunate dilemma: many domains where MHPs
would enable the most promising applications are
exactly those that are intractable with current inference methods.

In this work, we propose a new approach to overcome this issue in the context of
large-scale diffusion processes. Our approach is based on the important observation
that such processes exhibit a characteristic sparsity pattern in that only a
very small fraction of all possible entities participate in any given event
sequence. To exploit this property, we develop a combined model and inference
method that does not only allow to compute the \emph{exact} likelihood and gradients
on large-scale data, but also improves runtime performance by multiple orders of
magnitude -- even when compared to state-of-the-art neural methods. In our
experiments, we show that this enables interpretable models of diffusion
processes at previously unattainable scale.

The remainder of this paper is structured as follows: In \Cref{sec:related}, we discuss
related work. In \Cref{sec:model}, we discuss our model and how to train it at
scale. In \Cref{sec:experiments}, we evaluate our method both with regard to model
quality and runtime complexity on synthetic and real-world data.

\section{Related Work}
\label{sec:related}
Temporal point processes have long been explored for modeling diffusion
processes in social networks. Following the pioneering work of
\citet{gomez-rodriguez2011uncovering} on uncovering latent influence structures
from information cascades, TPPs have been applied, for instance, to modeling
information pathways in online media, \citet{gomez-rodriguez2013structure}, to
analyze the diffusion of policies in US states \cite{boehmke2018new}, and to topic
modeling from text cascades \cite{he2015hawkestopic}. However, none of these
methods can scale to a large number of entities that participate in the diffusion
process and simultaneously model their interdependencies. Other methods such as
SEISMIC \citep{zhao2015seismic} or CONTINEST \citep{du2013scalable} and COEVOLVE
\cite{farajtabar2017coevolve} model information cascades using either univariate
HPs, or require the underlying network structure to be known.
\citet{lemonnier2017multivariate} proposed factorized MHPs which improve runtime
and memory complexity significantly with respect to the number of nodes in an
influence network. However, even this approach requires more than \(3.7 \cdot
10^5\) seconds of training time for a relatively small dataset with \(1075\) nodes.
Hence, no existing method is currently able to model large-scale influence
structures even though there has been considerable work in applying TPPs and
MHPs to model diffusion processes over the years.

Another important approach that has recently received increased attention
is to model TPPs using recurrent neural architectures. This
includes, for instance Recurrent Marked Temporal Point Processes (RMTPP;
\citealt{du2016recurrent}), Dynamic Embeddings for Temporal Recommendations
\citep{kumar2019predicting}, and Neural Hawkes Processes (NHPs;
\citealt{tpp/mei2017neuralhawkes}). An appealing property of neural methods for
TPPs is that they scale well with regard to the length of an event sequence and
offer a flexible framework to parameterize an intensity function. However,
current neural methods do not scale well to datasets that consist of a large
number of entities \emph{and} a large number of event sequences. Moreover, most
existing neural methods are essentially black-box methods that are not
interpretable.

\begin{figure}[t]
\centering
\includegraphics[width=0.8\columnwidth]{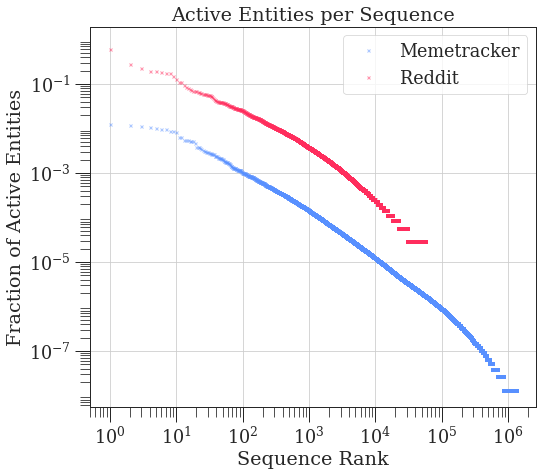}
\caption{\label{fig:active-dim}\small Sparsity of event sequences on Memetracker and Reddit Hyperlinks (log-log scale). Sequences are ranked by their fraction of active entities. For most sequences, only a minuscule portion of all possible entities is active.}
\end{figure}

Another shortcoming of many recurrent neural methods is that they need to resort
to uniform sampling to compute the compensator of a TPP. This is clearly
suboptimal for application areas with very irregular time intervals such as
bursty information cascades. For this reason, another line of research has
recently explored methods based on Neural ODEs \citep{chen2018neural} which allow
to compute the exact intensity function using neural methods. This includes, for
instance, Latent ODEs for irregularly sampled time series
\citep{rubanova2019latentodes} and Neural Jump Stochastic Differential Equations
\citep{jia2019neuraljump}. However, current Neural ODE-based methods suffer from
the same interpretability shortcomings as recurrent architectures and are also
difficult to scale and train. Another promising line of research explores
intensity-free learning of TPPs to circumvent some of the computational issues
associated with parameterizing intensity functions
\citep{shchur2019intensityfree,xiao2017wassersteinpp}.

In this work, we pursue a different approach for modeling large-scale diffusion
processes: by going back to the original idea of MHPs and by carefully adjusting
model and inference method to properties of the data, we show that MHPs can in
fact be applied to large-scale domains while retaining all of its appealing
properties including interpretabtiliy and state-of-the-art performance.

\section{Method}
\label{sec:method}
In the following, let \(t^x\) denote an event occuring at time \(t\) and at entity
\(x \in \Set{X}\).\footnote{For notational convenience, we will simply write \(t_i\) if the identity
of an event's entity is not relevant in a given context.}
Furthermore, let \(h = \{t_i^x\}_{i=1}^n\) denote a single event sequence with
\(t_i < t_{i+1}\). Furthermore, let \(T_h\) denote the maximum observation time for
event sequence \(h\). A temporal point process is then fully characterized through its \emph{conditional
intensity} function
\begin{equation}
    \lambda_x(t|h) = \lim_{\Delta t \downarrow 0 }\frac{\Pr\left( t_i^x \in [t, t+\Delta t)\ |\ h \right)}{\Delta t}
\end{equation}
which specifies the infinitesimal probability that an event \(t_i^x\) occurs in
the time interval \([t, t+\Delta t)\) given past events \(h\). We follow
\citet{daley2008introduction} and use \(\lambda^*_x(t)\) as shorthand for \(\lambda_x(t|h)\).
Given a dataset \(\hist = \{h_i\}_{i=1}^m\) of event sequences, we can then jointly
estimate the parameters for all \(\lambda_x\) by maximizing the log-likelihood
\begin{equation}
    \LL(\hist) = \sum_{h \in \hist} \sum_x \left(\sum_{i=1}^n \log \lambda^*_x(t_i) - \int_0^{T_h} \lambda^*_x(s)ds\right)
    \label{eq:ll}
\end{equation}

Prior work on scaling MHPs has mainly focused on the integral (compensator) in
\Cref{eq:ll} which is intractable to compute in general. However, a second
important aspect regarding scalability is the number of entities \(|\Set{X}|\) and
the number of event sequences \(|\hist|\): It can be seen from \Cref{eq:ll} that the
computation of the log-likelihood for a \emph{single} sequence \(h\) involves all
entities of the TPP. Hence, the overall runtime complexity for a
single evaluation of the log-likelihood (and its gradient) becomes \(O(|\Set{X}|
\cdot |\hist|)\).\footnote{In our runtime analysis, we absorb the cost
of \(\lambda_x(t|h)\) into \(|\hist|\) since it does not differ
between the model classes we consider.} This becomes prohibitively expensive in
domains such as social networks and information systems, where both \(|\Set{X}|\)
and \(|\hist|\) can easily be in the millions.

To overcome this issue, our approach is based on the important observation that
sequences in large-scale domains are typically \emph{sparse}, i.e., that only a
small fraction of all possible entities participate in a given episode. For
instance, only very few (out of all possible) users in a social network
typically interact with (e.g., like, share) a given meme. \Cref{fig:active-dim}
illustrates this property on two larger real-world datasets: \emph{Memetracker}
\citep{leskovec2009meme} which consists of \(\sim 75\)M meme cascades collected
from \({\sim1.3}\)M websites; and \emph{Reddit Hyperlinks} \citep{kumar2018community}
which consists of \({\sim860}\)K hyperlinks between \({\sim55}\)K Reddit
communities (subreddits). For Memetracker, it can be seen that even for the most
dense sequences, less than \(10\%\) of all entities are active and for the vast
majority of sequences it is even less than \(0.001\%\). The smaller Reddit
Hyperlinks dataset exhibits a similar distribution.
Based on these observations, we will now propose a combined model and
inference method to compute the exact likelihood and gradients at scale.

\paragraph{Model}
\label{sec:model}
To develop our model, we build on prior work on Exponential Hawkes Processes
which are defined through the conditional intensity function
\begin{equation}
    \lambda^*_x(t) = \mu_x + \sum_{t_i^y < t} \alpha_{xy} e^{-\beta(t - t_i^y)}
    \label{eq:ehp}
\end{equation}
where \(\mu_x, \alpha_{xy}, \beta > 0\) and \(t_i^y \in h\). For \Cref{eq:ehp}, it is
well known \citep{tpp/ozaki1979maximum} that the compensator can be computed in closed form via
\begin{equation*}
\int_0^{T_h} \lambda^*_x(s)ds = \mu_x T_h + \sum_{t_i^y \in h} \frac{\alpha_{xy}}{\beta}\left(1 - e^{-\beta(T_h - t^y_i)}\right)
\end{equation*}

Moreover, the Markovian nature of Exponential Hawkes Processes allows to compute
their log-likelihood in \(O(|h|)\), using a recursive algorithm \cite{ogata1981lewis}:
\begin{equation*}
    \LL(h) = \sum_x \sum_{i=1}^n \log \left(\mu_x + A(i)\right) - \int_0^{T_h} \lambda^*_x(s)ds
\end{equation*}
where
\begin{equation*}
    A(i) = \begin{cases}
         \sum_{t_j < t_i} \alpha_{xy} e^{-\beta(t_i - t^y_j)} & i > 1 \\
         0 & i = 1
    \end{cases}
\end{equation*}

However, while this solves some of the scaliability issues associated with MHPs,
it does not address the core concern of our work, i.e., scalability with regard to
\(|\Set{X}|\) and \(|\hist|\), as \Cref{eq:ll} would still have runtime complexity
\({O(|\Set{X}| \cdot |\hist|)}\) and memory complexity
\(O(|\Set{X}|^2)\) (due to \(\alpha\)).

Instead, we parameterize \Cref{eq:ehp} as follows: Let \(\param^\mu_x\),
\(\param^\beta\), \(\param^s_x\), \(\vpar^u_x \in \R^{d}\), and \(\vpar^v_y \in \R^{m}\)
be real-valued scalars and vectors, respectively. Furthermore, let \({\phi: \R
\to \R_+}\) be a positive differentiable function such as the Softplus function
\(\phi(\theta) = \log(1 + \exp(\theta))\). We then set
\begin{align*}
    \mu_{x} & = \phi(\param^\mu_x) &
    \beta_{\phantom{x}} & = \phi(\param^\beta) \\
    \vu_{x} & = \phi(\vpar^u_x) &
    \vv_{y} & = \phi(\vpar^v_y)
\end{align*}
and compute the influence parameters \(\alpha\) using the non-negative factorization
\begin{equation*}
  \alpha_{xy} =  \begin{cases}
    \phi(\param^s_x) & \text{if } x = y \\
    \vu_x^\top \vv_y & \text{otherwise}
  \end{cases}
\end{equation*}

This parameterization ensures that \(\lambda\)\textsubscript{x} stays always positive while simultaneously
being differentiable such that we can employ gradient-based optimization.

A crucial difference to prior work on Exponential Hawkes Processes is that we factorize the
influence structure as \({\alpha_{xy} = \langle \phi(\vpar^u_x),
\phi(\vpar^v_y)\rangle}\) as opposed to \({\alpha_{xy} = \phi(\langle \vpar^u_x,
\vpar^v_y \rangle)}\). This non-negative factorization is key to our scalable
inference method as we will show in the following.

\paragraph{Scalabale Likelihood and Gradient Computation}
\label{sec:gradients}
First, note that the log-likelihood can be decomposed into active
and inactive entities in an event sequence:
\begin{equation}
    \LL(\hist) = \sum_{h \in \hist} \left(\sum_{x \in h} \LL_x(h) + \sum_{x \not \in h} \LL_x(h)\right)
\end{equation}
Second, for inactive entities \(x \not \in h\), \(\LL_x\) simplifies to
\begin{equation*}
    \forall x \not \in h: \LL_x(h) = \int_0^{T_h} \lambda_x^*(s)ds
\end{equation*}
Furthermore, due to our non-negative factorization, \(\alpha_{xy}\) is now a linear function of
\(\vu_x\) and \(\vv_y\) and we can therefore rewrite
\(\LL_x(h)\) for inactive entities as
\begin{multline*}
    \sum_{x \not \in h}\LL_x(h) =
        -\sum_{x\not \in h} \mu_x T_h \\ - \frac{1}{\beta} \Bigg(\sum_{x \not \in h} \vu_x\Bigg)^\top\sum_{t_i^y \in h} \vv_y \left(1 - e^{-\beta(T_h - t^y_i)}\right)
\end{multline*}

Combining these three properties, our approach is then based on the following
idea: for sparse event sequences
where \(|x \in h| \ll |x \not \in h|\) it holds in terms of runtime complexity that
\begin{equation*}
O\Bigg(\sum_{x \not \in h} \vu_x \Bigg) \ll O\Bigg(\widehat{\vu} -
\sum_{x \in h}\vu_x\Bigg)
\end{equation*}
where \(\widehat{\vu} = \sum_x \vu_x\). This is the case because \(\widehat{\vu}\) is
identical for all \(x \in \Set{X}\) and \(h \in \hist\). We can therefore regard
it as a constant that has to be computed only once.

Using this reparameterization trick, we can then compute the full likelihood via
\begin{align}
\LL(\hist) & = \sum_{h \in \hist} \left(\sum_{x \in h} \LL_x(h) - \frac{1}{\beta}C_x - \mu_x D_x \right) \label{eq:ll-lazy} \\
    C_x & = \left(\widehat{\vu}_h - \vu_x\right)^\top \sum_{t^y_i \in h} \vv_y \left(1 - e^{-\beta (T_h - t_i^y)}\right) \notag \\
    D_x & = \sum_{h \in \hist_x^-} T_h\,/\, |\hist_x^+| \notag
\end{align}
where \(\widehat{\vu}_h = \widehat{\vu} / (|x \in h|)\) and where
\(\hist_x^+ = \{h : x \in h\}\) and \(\hist_x^- = \{h : x \not \in h\}\)
are the sets of all sequences in which \(x\) does and does not occur,
respectively. Although \(D_x\) depends on \(\hist_x^-\), it depends on no parameter
of the model. It can therefore be again precomputed and regarded as a constant.
Furthermore, let \(E\) denote the average number of
active entities per event sequence. It can then easily be verified that
\Cref{eq:ll-lazy} has a runtime complexity of \({O(|\Set{X}| + |\hist| \cdot
E)}\) since processing each sequence \(h\) depends only on precomputed
quantities for entities \({x \not \in h}\). For sparse sequences where \(E
\ll |\Set{X}|\), \Cref{eq:ll-lazy} can therefore yield substantial runtime improvements
compared to existing inference methods.

To compute the gradients of \Cref{eq:ll}, we can apply similar ideas. In
particular, the gradient for post-activation weights \(\mu_x\) can be computed as
\begin{align}
\frac{\partial}{\partial \mu_x}\LL(\hist) & = \sum_{h \in \hist_x^+} \left(\frac{\partial}{\partial \mu_x}\LL_x(h) - G^\mu_x \right) \label{eq:grad-mu}\\
G^\mu_x & = \sum_{h \in \hist_x^-} T_h\, /\, |\hist_x^+|\ . \notag \\
\intertext{For $\beta$, the gradient can be computed via}
\frac{\partial}{\partial \beta}\LL(\hist) & = \sum_{h \in \hist} \sum_{x \in h} \left(\frac{\partial}{\partial \beta} \LL_x(h) - G_x^\beta \right) \label{eq:grad-beta}\\
%G_x^\beta & = \left(\widehat{\vu}_h - \vu_x\right)^\top \sum_{t_i^y \in h} \vv_y(T_h - t_i^y)e^{-\beta(T_h - t_i^y)} \notag \\
G_x^\beta & = \left(\widehat{\vu}_h - \vu_x\right)^\top \sum_{t_i^y \in h} \vv_y \frac{\partial}{\partial \beta}\frac{1}{\beta}\left(1 - e^{-\beta(T_h - t_i^y)}\right) \notag \\
\intertext{For $\vu_x$, the gradient can be computed via}
\frac{\partial}{\partial \vu_x}\LL(\hist) & = \sum_{h \in \hist_x^+} \left(\frac{\partial}{\partial \vu_x} \LL_x(h) + \frac{1}{\beta}G^{\vu}_x\right) \label{eq:grad-u}\\
G^{\vu}_x & = \sum_{t_i^y \in h} \vv_y \left(1 - e^{-\beta(T - t_i^y)}\right) - \widehat{\vec{z}}\notag \\
\intertext{where $\widehat{\vec{z}} = \sum_{h}\sum_{t_i^y \in h}  \vv_y \left(1 - e^{-\beta(T_h - t_i^y)}\right)$. Similar to $\widehat{\vu}$, $\widehat{\vec{z}}$ is identical for all $x$ and $h$ and, therefore, needs to be computed only once per epoch.}
\intertext{For $\vv_y$ the gradient can be computed via}
\frac{\partial}{\partial \vv_y}\LL(\hist) & = \sum_{h \in \hist} \left(\sum_{x \in h} \frac{\partial}{\partial \vv_y} \LL_x(h) - \frac{1}{\beta}G^{\vv}_y \right) \label{eq:grad-v}\\
G^{\vv}_y & = \left(\widehat{\vu} - \vu_x \right) \sum_{t_i = y} \left(1 - e^{-\beta(T_h - t_i^y)}\right) \notag
\end{align}

Finally, self-excitation parameters \(\alpha_{xx}\) receive gradients only for
sequences where \(x \in h\) and require no special treatment.
For the full derivation of all calculations, we refer
the reader to the supplementary material.

By using this approach, we can now compute the exact log-likelihood and
gradients of the model by processing only the active entities in an event
sequence and with an overall runtime complexity of \({O(|\Set{X}| + |\hist| \cdot
E)}\). We will refer to this combined model and inference method as \emph{Lazy
MHPs}.

\paragraph{Stochastic Training}
\label{sec:stochastic}
The likelihood and gradients in \Cref{sec:gradients} can directly be used for
parameter estimation using batch methods such as L-BFGS \cite{byrd1995limited}.
However, batch optimization can be difficult to scale and sensitive to
initialization and local minima. For this reason, we discuss in the following how
to apply stochastic optimization methods in our setting.

Naive stochastic optimization using the gradients in
\Cref{sec:gradients} would again lead to bad runtime performance as it would
require to compute \(\widehat{\vu} = \sum_x \vu_x\) for each minibatch.
However, due to is linear structure, we can update \(\widehat{\vu}\)
on-the-fly during the optimization process. In particular, let \(\vpar^{u-}_x\),
\(\vpar^{u+}_x\) denote parameters before and after an update. We
can then simply compute the new state of \(\widehat{\vu}\) via
\begin{equation*}
    \widehat{\vu} \gets \widehat{\vu} - \phi(\vpar^{u-}_x) + \phi(\vpar^{u+}_x)
\end{equation*}
As such, it is sufficient to compute \(\widehat{\vu}\) only once prior to training. The
second is \(\widehat{\vec{z}}\). Since \(\widehat{\vec{z}}\) is not a linear
function of either \(\vv_y\) or \(\beta\), we employ a different strategy: First,
note that \(\widehat{\vec{z}}\) can be computed in one pass through the data and
by only considering active entities. Hence, we can compute \(\widehat{\vec{z}}\)
on-the-fly while processing an epoch and use it in the next epoch to
compute gradients. While this leads to lagging updates for \(\vu_x\), we did not
observe any deterioration in terms of convergence or model quality. The full
algorithm for stochastic training of Lazy MHPs is listed in \Cref{alg:sgd}.

\begin{algorithm}[t]
\caption{Stochastic Training of Lazy MHPs\label{alg:sgd}}
\begin{algorithmic}
  \STATE $\widehat{\vu} \gets \sum_x \vu_x$
  \STATE $\vs{0} \gets \sum_{h \in \hist} \sum_{t_i^y \in h} \vv_y \left(1 - e^{-\beta(T_h - t_i^y)}\right)$
  \STATE $\forall x: C^\mu_x \gets$ using \Cref{eq:grad-mu}
  \item[]

  \FOR{each epoch $\ell$}
  \FOR{$h \in \hist$}
    \STATE $\vs{\ell} \gets \vs{\ell} + \sum_{t_i^y \in h} \vv_y \left(1 - e^{-\beta(T_h - t_i^y)}\right)$
    \FOR{$x \in h$}
      \STATE $\nabla\param^\mu_x \gets$ using \Cref{eq:grad-mu} and $C^\mu_x$
      \STATE $\nabla\param^\beta \gets$ using \Cref{eq:grad-beta} and $\widehat{\vu}$
      \STATE $\nabla\vpar^u_x \gets$ using \Cref{eq:grad-u} and $\vec{z}^{\ell - 1}$
      \STATE $\nabla\vpar^v_y \gets$ using \Cref{eq:grad-v} and $\widehat{\vu}$
      \STATE $\Theta \gets$ Stochastic parameter update
      \STATE $\widehat{\vu} \gets \widehat{\vu} - \phi(\vpar^{u-}_x) + \phi(\vpar^{u+}_x)$
    \ENDFOR
  \ENDFOR
  \ENDFOR
\end{algorithmic}
\end{algorithm}

\paragraph{Implementation and Training Details}
\label{sec:org122419c}
Since the efficient computation of recursive gradients is not supported by
standard frameworks such as PyTorch or Tensorflow, we implemented our model as a
custom C++ extension to PyTorch. In our implementation, we exploit that a large
number of sequences and entities allows for massively parallel training using
Hogwild \cite{DBLP:conf/nips/RechtRWN11} as we can dispatch each \(\LL_x(h)\) to a
separate thread. This has the advantage over mini-batching that we do not need
to resort to padding or to group sequences by length (both would be problematic
as sequence length can differ by multiple orders of magnitude). For
optimization we employ Adam \cite{kingma2014adam}.

\paragraph{Scope and Limitations}
\label{sec:scope}
Our approach is explicitly designed to train MHPs in large-scale domains with
possibly millions of entities and sparse event sequences. In smaller domains or
on data with dense sequences, our approach will likely not provide significant
advantages over traditional methods. Furthermore, our method relies on a
non-negative factorization of \(\alpha\) to achieve scalability. While this has
the additional positive effect of interpretable latent factors, it will also
lead to larger embedding dimensions compared to standard factorizations, which
can have negative effects on runtime performance. Due to the non-negative
factorization, our method is also limited to modeling mutual excitation (e.g., it
cannot model inhibition). Overall, we believe these are acceptable
tradeoffs for learning large-scale MHPs.

\section{Experiments}
\label{sec:experiments}
In the following, we evaluate our model on various synthetic and real-world
datasets to demonstrate three key properties of our method: a) Lazy MHPs improve
runtime by multiple orders of magnitude compared to existing methods b) since
Lazy MHPs compute exact gradients, we are able to infer accurate model
parameters c) interpretable model parameters allow us to analyze diffusion
processes at scale.

\begin{figure*}[t]
\centering
\begin{minipage}{\textwidth}
\begin{subfigure}{.25\textwidth}
    \includegraphics[width=\columnwidth]{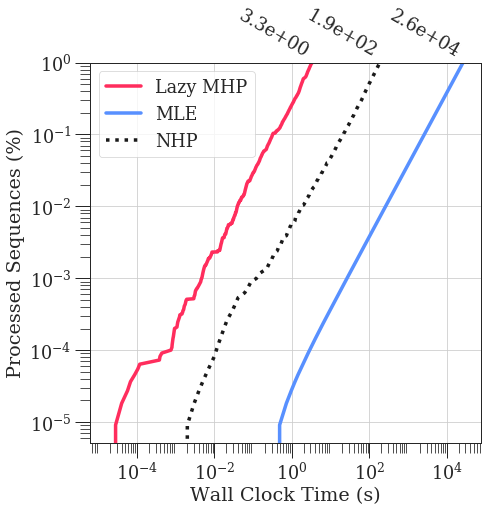}

    \vspace{1ex}

    \includegraphics[width=\columnwidth]{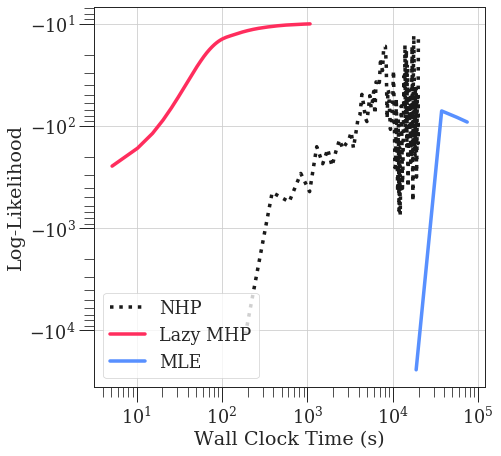}
    \caption{Amy Winehouse}
\end{subfigure}%
\begin{subfigure}{.24\textwidth}
    \includegraphics[width=\columnwidth]{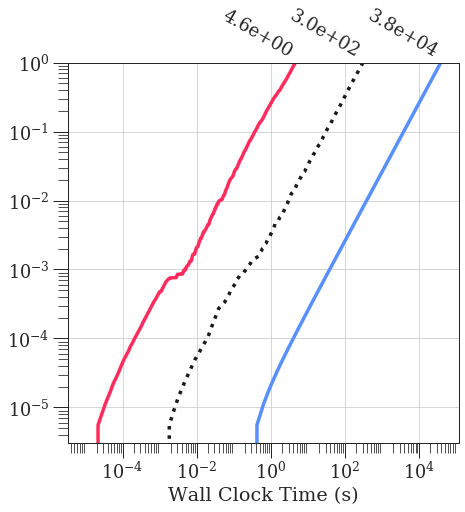}

    \vspace{1ex}

    \includegraphics[width=\columnwidth]{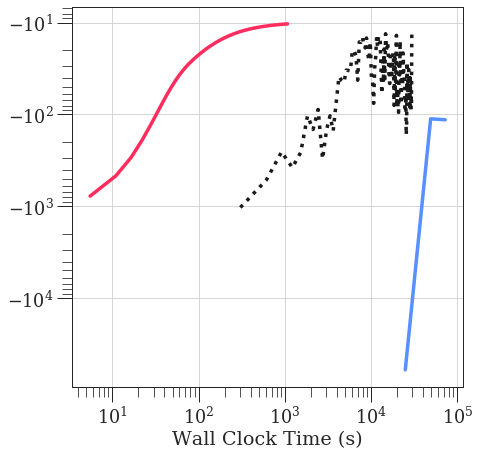}
    \caption{Arab Spring}
\end{subfigure}%
\begin{subfigure}{.24\textwidth}
    \includegraphics[width=\columnwidth]{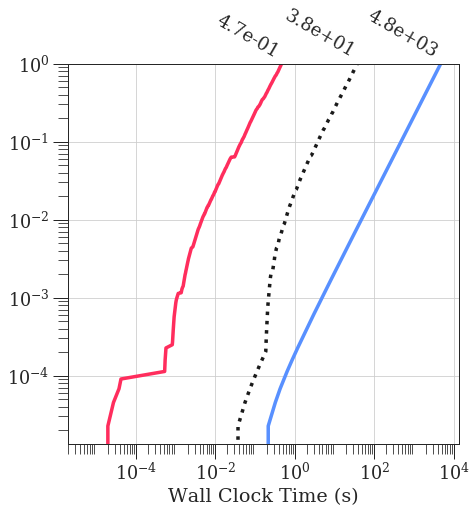}

    \vspace{1ex}

    \includegraphics[width=\columnwidth]{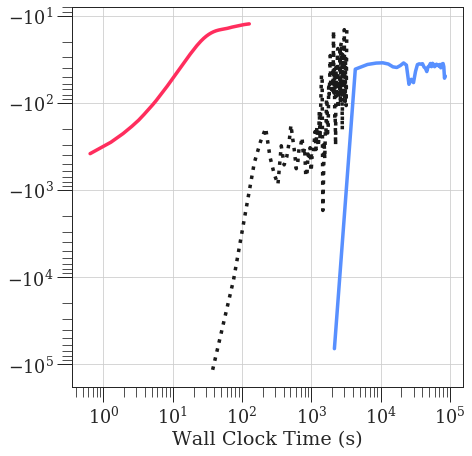}
    \caption{Bail out}
\end{subfigure}%
\begin{subfigure}{.24\textwidth}
    \includegraphics[width=\columnwidth]{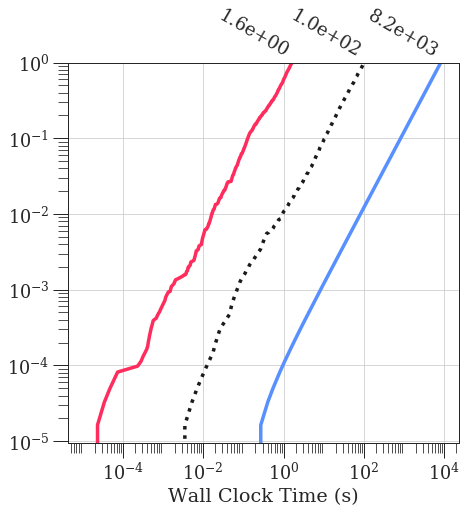}

    \vspace{1ex}

    \includegraphics[width=\columnwidth]{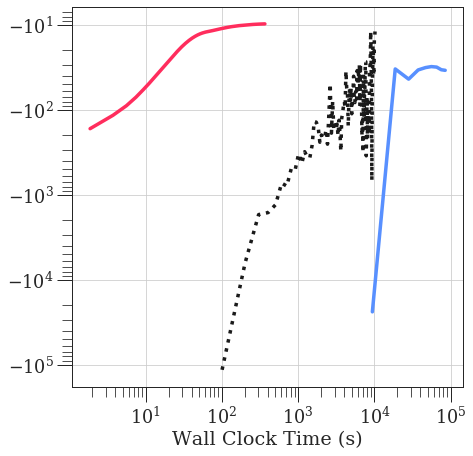}
    \caption{Miami Heat}
\end{subfigure}
\caption{Runtime experiments on Memetracker subsets (log-log scale).\label{fig:memetracker-runtime}}
\end{minipage}
\end{figure*}

We compare Lazy MHPs to typical factorized MHPs where \({\alpha_{xy} =
\phi(\langle \vpar^u_x, \vpar^v_y \rangle)}\) and which are trained with standard
maximum-likelihood (MLE). This allows us to compare both the performance of lazy
training as well as the effect of our non-negative factorization. If applicable,
we also compare to ADM4 \citep{zhou2013learning} as implemented in \emph{tick}
\citep{bacry2017tick}. Furthermore, we compare to Neural Hawkes Processes
\citep{tpp/mei2017neuralhawkes} as a scalable neural method. Due to space
constraints, we report full hyperparameter settings in the supplementary material.

\begin{table}[b]\small
\caption{\label{tab:memetracker-stats}\small\vspace{-1.5ex} Memetracker Topic Statistics}
\centering
\begin{tabular}{lrrr}
\toprule
 & \textbf{Nodes} & \textbf{Episodes} & \textbf{Events}\\
\midrule
Amy Winehouse & 1,561 & 109,650 & 226,247\\
Arab Spring & 1,377 & 179,681 & 400,199\\
Bail Out & 835 & 44,130 & 64,138\\
Miami Heat & 1,173 & 61,280 & 133,451\\
\bottomrule
\end{tabular}
\end{table}

\paragraph{Runtime Analysis on Real-World Cascades}
\label{sec:org450712d}
First, we focus on the runtime performance of our model on real-world
cascades. For this purpose, we employ various subsets of \emph{Memetracker}
\citep{leskovec2009meme} as extracted by \citet{gomez-rodriguez2013structure}. Each
subset consists of information cascades that are related to a specific topic in
the news cycle from March 2011 to February 2012. For comparison, we selected
four subsets whose sizes are still tractable with standard MHPs (see
\Cref{tab:memetracker-stats} for data statistics).
Unfortunately, these datasets were already too large for ADM4 and EM as
implemented in \emph{tick}, causing out-of-memory errors on 500GB memory machines.

\begin{figure*}[t]
\begin{minipage}{\textwidth}
\captionof{table}{\small Model quality on synthetic data and example of generated sequences.\label{tab:synth-rmse}}
\begin{minipage}{.8\textwidth}
\centering
\small
\vspace{-1ex}
\begin{tabular}{llcccccccc}
\toprule
& & & & \multicolumn{6}{c}{\bf \footnotesize RMSE} \\
\cmidrule(lr){5-10}
& & \multicolumn{2}{c}{\bf Log-Likelihood} & \multicolumn{2}{c}{$\mu$} & \multicolumn{2}{c}{$\beta$} & \multicolumn{2}{c}{$\alpha$} \\
\cmidrule(lr){3-4} \cmidrule(lr){5-6} \cmidrule(lr){7-8} \cmidrule(lr){9-10}
& & 50 & 250 & 50 & 250 & 50 & 250 & 50 & 250 \\
\cmidrule(lr){1-2}\cmidrule(lr){3-4} \cmidrule(lr){5-6} \cmidrule(lr){7-8} \cmidrule(lr){9-10}
\multirow{4}{*}{\bf Low Rank} & ADM4 & -6.88 & -8.40 & 2.8e-5 & 1.0e-5 & - & - & 0.009 & 0.003  \\
& NHP & -7.34 & -8.93 & - & - & - & - & - & - \\
& MLE & -6.93 & -8.60 & 3.0e-5 & 1.2e-5 & 0.017 & 0.006 & 0.009 & 0.003 \\
& Lazy MHP & -6.96 & -8.64 & 2.8e-5 & 1.2e-5 & 0.001 & 0.038 & 0.015 & 0.007 \\
\cmidrule(lr){1-2}\cmidrule(lr){3-4} \cmidrule(lr){5-6} \cmidrule(lr){7-8} \cmidrule(lr){9-10}
\multirow{4}{*}{\bf Full Rank} & ADM4 & -7.27 & -8.50 & 2.6e-5 & 1.1e-5 & - & - & 0.011 & 0.003 \\
& NHP & -8.02 & -9.13 & - & - & - & - & - & - \\
& MLE & -7.28 & -8.62 & 2.8e-5 & 1.3e-5 & 0.022 & 0.060 & 0.015 & 0.004 \\
& Lazy MHP & -7.28 & -8.61 & 2.7e-5 & 1.1e-5 & 0.015 & 0.037 & 0.037 & 0.008 \\

\bottomrule
\end{tabular}
\end{minipage}%
\begin{minipage}{.2\textwidth}
\centering
\includegraphics[angle=90,origin=c, width=.8\columnwidth]{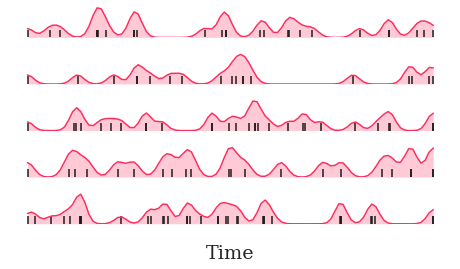}
\end{minipage}
\end{minipage}
\begin{minipage}{\textwidth}
\centering
\begin{minipage}{.55\textwidth}
\begin{subfigure}{\columnwidth}
\includegraphics[width=\columnwidth]{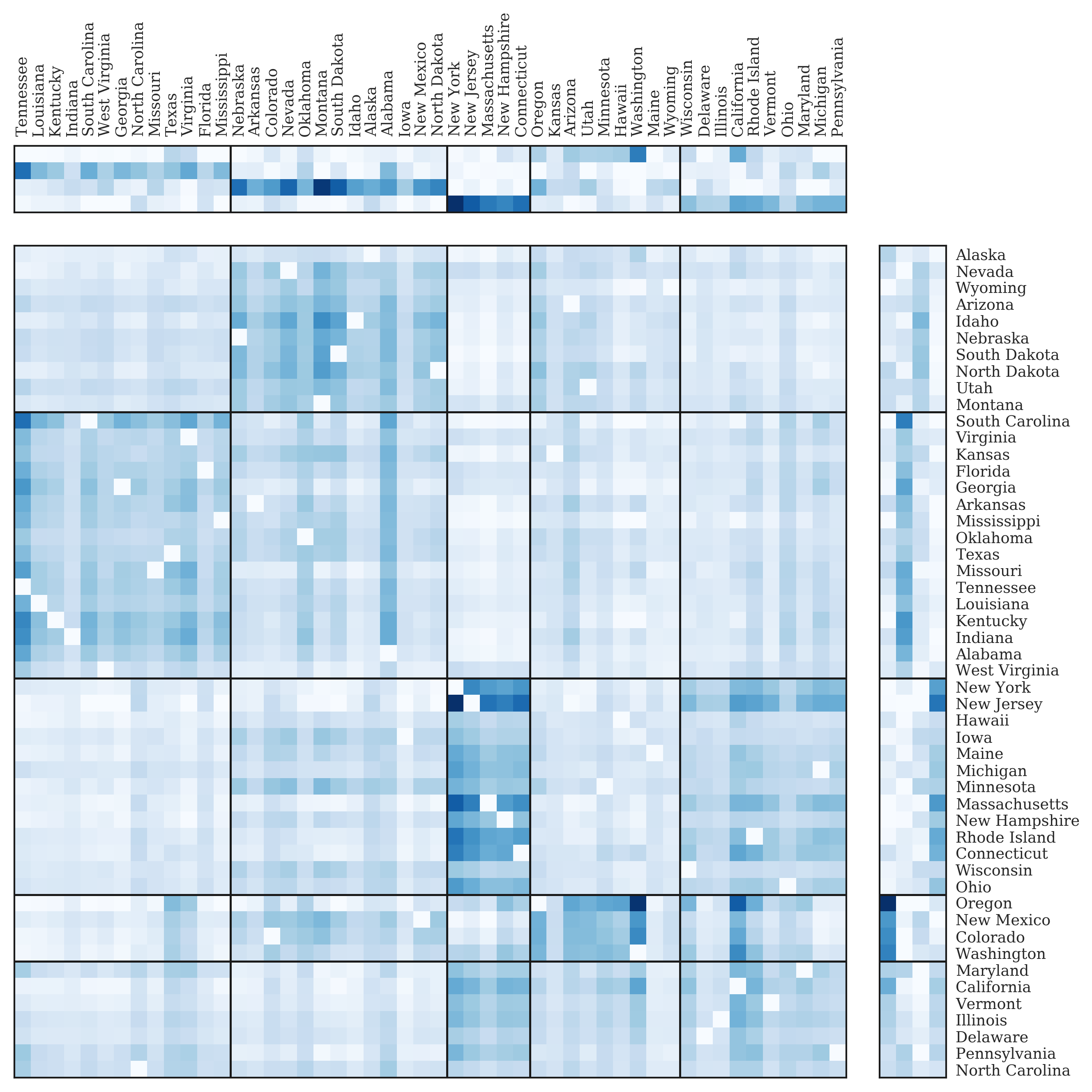}
\end{subfigure}
\end{minipage}%
\begin{minipage}{.4\textwidth}
    \vspace{1em}

    \begin{subfigure}{.9\columnwidth}
    \includegraphics[width=\columnwidth,trim=100 50 100 100,clip]{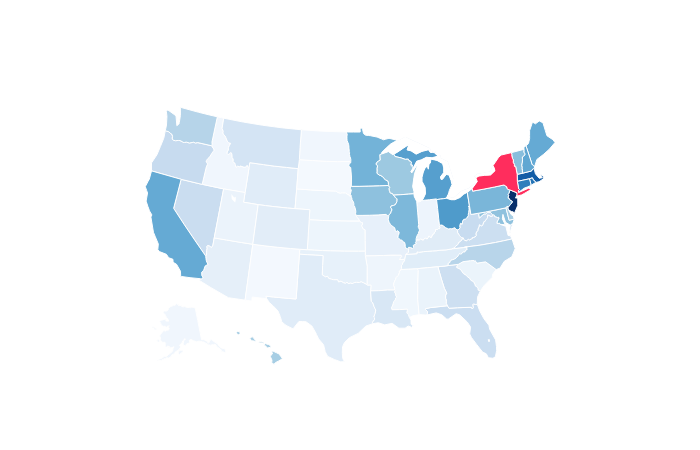}
    \caption{Influence of New York\label{fig:spid-ny}}
    \end{subfigure}

    \vspace{1em}

    \begin{subfigure}{.9\columnwidth}
    \includegraphics[width=\columnwidth,trim=100 50 100 100,clip]{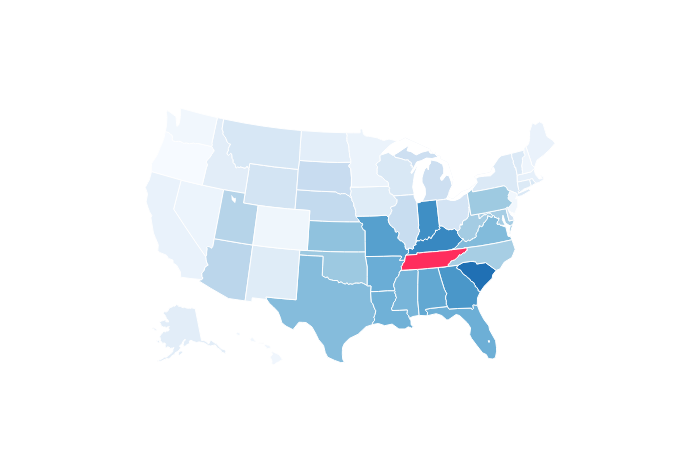}
    \caption{Influence of Tennessee\label{fig:spid-tn}}
    \end{subfigure}
\end{minipage}
\captionof{figure}{\small Latent structure of policy diffusion in US states.\label{fig:spid}}
\end{minipage}
\end{figure*}

In our experiments, we compared the runtime for a single epoch on these datasets
for: Lazy MHPs, standard MHPs (MLE), and the Neural Hawkes Process (NHP). Both MHP
models were trained using 40 threads while the NHP was trained
on a GPU. It can be seen from \Cref{fig:memetracker-runtime} (upper row)
that Lazy MHPs improve the runtime compared to standard MHPs by approximately
\emph{four orders of magnitude} on all datasets. Even for Neural Hawkes
Processes -- which are one of the most scalable available methods -- our method
improves the runtime by approximately \emph{two orders of magnitude}. This is
remarkable as NHPs are a black-box model that makes various approximations to
acheive scalability while Lazy MHPs compute an exact, interpretable MHP. When
comparing methods in terms of log-likelihood, we see similar results
(\Cref{fig:memetracker-runtime} lower row). Lazy MHPs converge dramatically faster
and to substantially better log-likelihoods compared to both MLE and NHP and on
all datasets (we limited runtime to 24 hours). It can also be seen that the
training of NHPs is substantially less stable on these datasets

\paragraph{Parameter Estimation on Synthetic Data}
\label{sec:org5aeb337}
Next, we evaluate our model on synthetic datasets that have been generated from
standard MHPs with known parameters \(\mu\), \(\beta\), and \(\alpha\). This allows us
to not only test our stochastic training method, but also evaluate the ability
of our non-negative factorization to recover the correct model parameters
\(\alpha\). In detail, the datasets have been generated as follows: We first
sampled random directed scale-free graphs as proposed by
\citet{bollobas2003directed} to generate synthetic influence structures. We varied
the size of the graph between 50 and 250 nodes. Furthermore, we set the global
timescale parameter \(\beta = 1.0\) and base intensities \(\mu = 0.0001\). We then
sampled \(1000\) event sequences from these MHPs using thinning in two
variants: 1) where we use the original full-rank adjacency matrix of the
influence graphs and 2) where we us a 10-dimensional low-rank approximation of
it. Both settings resulted in sparse and bursty event sequences (see
\Cref{tab:synth-rmse} for examples of generated sequences).

In \Cref{tab:synth-rmse}, we report the RMSE of the inferred parameters as well as
the log-likehood of the observed sequences. Since ADM4 requires \(\beta\) as a
hyperparameter, we provide it with the ground-truth value during training. As
such ADM4 has a distinct advantage in our evaluation. For the Lazy MHP and MLE
models we used a maximum factorization dimension of \(d=20\). It can be seen that
Lazy MHPs show strong results in this task and accurately recover the model
parameters as compared to ADM4 and MLE. The log-likelihood results of Lazy MHPs
show similarly strong results. The non-negative factorization has only a limited
effect on model quality when compared to MLE. Furthermore, even for the
full-rank setting a \(d=20\) factorization was sufficient to achieve
strong results. All MHP based methods outperform NHPs on these datasets, which is
likely due to the mismatch of bursty event sequences and uniform sampling in
NHPs (see suppl. material for a detailed evaluation of this aspect)
\vspace{-1ex}

\begin{table*}[t]
\small
\centering
\caption{Latent factors of Reddit Hyperlink model.\label{tab:reddit}}
\adjustbox{max width=\textwidth}{
\begin{tabular}{llll}
\toprule
\bf Topic & \bf Subreddits & \bf Topic & \bf Subreddits \\
\cmidrule(lr){1-2}\cmidrule(lr){3-4}
\it Music &  music, hiphopheads, metal, indieheads, metaljerk &
\it Football & nfl, greenbaypackers, seahawks, panthers, detroitlions \\
& electricforest, electronicmusic, punk, listentothis &
& falcons, kansascitychiefs, steelers, miamidolphins \\
\cmidrule(lr){1-2}\cmidrule(lr){3-4}
\it Music Prod. & wearethemusicmakers, edmproduction, bestof &
\it Basketball & nba, bostonceltics, atlantahawks, sixers, nbaspurs \\
& monstercat, audioengineering, djs, music, synthesizers &
& warriors, pacers, chicagobulls, lakers, denvernuggets \\
\cmidrule(lr){1-2}\cmidrule(lr){3-4}
\it Movies & movies, starwars, marvel, arrow, marvelstudios, &
%\it E-Cigarettes & electronic\_cigarette, ecigclassifieds, ecr\_eu, &
\it Baseball & baseball, sfgiants, kcroyals, braves, azdiamondbacks \\
& startrek, fantheories, comicbooks, titlegore, dccomics &
%& rba, canadian\_ecigarette, vaping101, diy\_ejuice &
& phillies, keepwriting, minnesotatwins, padres, brewers \\
\cmidrule(lr){1-2}\cmidrule(lr){3-4}
\it Fashion & malefashionadvice, diy, streetwear, frugalmalefashion &
\it Role Playing & dnd, rpg, worldbuilding, lfg, dndbehindthescreen \\
& sneakers, charlieputh, watches, octobersveryown  &
& dndnext, boardgames, gameofthrones, roleplay \\
\cmidrule(lr){1-2}\cmidrule(lr){3-4}
\it Computers & techsupport, linux, sysadmin, buildapc, windows10 &
\it Pokemon & pokemon, postpreview, pokemongo, test, thesilphroad \\
& linuxquestions, linux4noobs, datahoarder, titlegore &
& pokemongiveaway, casualpokemontrades, pokemonplaza \\
\cmidrule(lr){1-2}\cmidrule(lr){3-4}
\it Electronics & arduino, electricians, askelectronics, askengineers &
\it Elite &  elitedangerous, aislingduval, kumocrew, elitemahon \\
& raspberry\_pi, electronics, thebutton, homeimprovement &
& elitepatreus, elitewinters, elitetorval, elitelavigny \\
\bottomrule
\end{tabular}
}
\end{table*}

\paragraph{SPID}
\label{sec:org0996cf1}
In the following, we examine the interpretability of our model with respect to
the inferred influence structure and latent factors. For this purpose, we employ
the \emph{State Policy Innovation and Diffusion Database (SPID;
\citealt{boehmke2019spid_data})} which has been collected to study innovations in
public policy and the spread of policies across US states. The database consists
of 728 policies established between the years 1691 to 2017. For each policy, the
SPID records the year of its eventual adoption in a state (e.g., \emph{(Alcoholic
Beverage Control, 1926, Pennsylvania), (Concealed Carry, 1975, Alabama)}). We
consider each policy as a separate event sequence and embed the entire
database using a 4-dimensional Lazy MHP. The influence structure \(\alpha_{xy}\)
in the MHP reflects the increased likelihood that a state \(x\) adopts a
policy if state \(y\) has adopted the same policy earlier.

Since SPID is a smaller database, it allows us to visually inspect and analyze
the inferred model in its entirety. \Cref{fig:spid} shows the full influence matrix
\(\alpha\) as well as the inferred latent factors \(\vu_x\), \(\vv_y\) for each state.
To aid visual analysis, we also clustered both latent factors using k-means. For
the full influence matrix, we computed a bi-clustering based on the k-means
clusters of the latent factors. It can be seen that the inferred model reveals
interesting structures of the diffusion process: First, \(\alpha\) as well as
\(\vu_x\) and \(\vv_y\) exhibit clear regional structures. For instance, New York is
influential for its direct neighbors (esp. New Jersey, Massachusetts) and
north-eastern states in general. Tennessee on the other hand is mainly
influential for states in the south-east such as South Carolina and Mississippi;
while California is influential for states such
as Oregon and Washington. However, not all influence is purely regional. For
instance, although on different coasts, New York is also influential
for California, since both states share similar policy adoptions. State-specific
plots in \Cref{fig:spid} further illustrate these patterns on the examples of New
York and Tennessee. Overall, New York and California are two of the most
influential states in our model, what matches previous analyses
\cite{boehmke2018new}.

\paragraph{Reddit Hyperlinks}
\label{sec:org4426185}
At last, we apply our model also on a larger-scale dataset, i.e., \emph{Reddit
Hyperlinks} \citep{kumar2018community} which consists of timestamped hyperlinks
between Reddit communities. In total, the dataset consists of over \(55\)K
entities (subreddits) and \(860\)K hyperlinks between them. We consider each
link target as a separate sequence and embed the entire dataset using a
50-dimensional Lazy MHP. The influence structure \(\alpha_{xy}\) reflects then the
increased likelihood that a subreddit \(x\) links to a target if subreddit \(y\) has
linked to it earlier.

Due to the non-negative factorization of Lazy MHPs, it is relatively easy to
analyze such larger datasets. Since we can interpret each value \(\vu_{xk}\) (and
\(\vv_{xk}\)) as the participation of entity \(x\) in component (or cluster) \(k\), we
can directly inspect the latent factors to analyze the latent structure of the
diffusion process. In \Cref{tab:reddit}, we list subreddits with the highest
activations (as measured by \(\vu_{xk}\)) for various latent
factors that have been inferred by our model. It can be seen that the model is
able to reveal topically consistent features that explain the temporal
interdependencies within hyperlink cascades -- ranging from relatively broad
topics such as music and sports to more focused communities about specific games
and electronics. Training a Lazy MHP requires \(\sim650\)s per epoch on this
dataset and achieves a log-likelihood of ca. \(-9.1\) after 100 epochs. To the best
of our knowledge, this is the first time that a MHP has been successfully
applied to a dataset of this size.

\section{Conclusion}
\label{sec:conclusion}
In this work, we developed a new method for modeling diffusion processes at
previously unattainable scale. By revisiting the original idea of MHPs and by
carefully adjusting model and inference to the sparsity of real-world
event sequences, we show that MHPs can in fact be applied to large-scale domains
while retaining all of its appealing properties including interpretabtiliy and
state-of-the-art performance. Our approach reduces runtime complexity from
\(O(|\Set{X}|\cdot|\hist|)\) to \(O(|\Set{X}| + |\hist| \cdot E)\) what improves
runtime performance by multiple orders of magnitude on sparse sequences. Since
scalability was an important limitation in existing methods, we believe that our
approach opens up promising new opportunities for modeling and understanding
social information systems at scale.

\bibliographystyle{icml2020}
\bibliography{paper}

\begin{thebibliography}{41}
\providecommand{\natexlab}[1]{#1}
\providecommand{\url}[1]{\texttt{#1}}
\expandafter\ifx\csname urlstyle\endcsname\relax
  \providecommand{\doi}[1]{doi: #1}\else
  \providecommand{\doi}{doi: \begingroup \urlstyle{rm}\Url}\fi

\bibitem[Achab et~al.(2017)Achab, Bacry, Ga{\"\i}ffas, Mastromatteo, and
  Muzy]{achab2017uncovering}
Achab, M., Bacry, E., Ga{\"\i}ffas, S., Mastromatteo, I., and Muzy, J.-F.
\newblock Uncovering causality from multivariate {Hawkes} integrated cumulants.
\newblock \emph{The Journal of Machine Learning Research}, 18\penalty0
  (1):\penalty0 6998--7025, 2017.

\bibitem[Alaa et~al.(2017)Alaa, Hu, and van~der Schaar]{alaa2017learning}
Alaa, A.~M., Hu, S., and van~der Schaar, M.
\newblock Learning from clinical judgments: Semi-{Markov}-modulated marked
  {Hawkes} processes for risk prognosis.
\newblock In \emph{Proceedings of the 34th International Conference on Machine
  Learning-Volume 70}, pp.\  60--69. JMLR. org, 2017.

\bibitem[Bacry et~al.(2015)Bacry, Mastromatteo, and Muzy]{bacry2015hawkes}
Bacry, E., Mastromatteo, I., and Muzy, J.-F.
\newblock Hawkes processes in finance.
\newblock \emph{Market Microstructure and Liquidity}, 1\penalty0 (01):\penalty0
  1550005, 2015.

\bibitem[{Bacry} et~al.(2017){Bacry}, {Bompaire}, {Ga{\"i}ffas}, and
  {Poulsen}]{bacry2017tick}
{Bacry}, E., {Bompaire}, M., {Ga{\"i}ffas}, S., and {Poulsen}, S.
\newblock {tick: a Python library for statistical learning, with a particular
  emphasis on time-dependent modeling}.
\newblock \emph{ArXiv e-prints}, July 2017.

\bibitem[Boehmke et~al.(2019)Boehmke, Brockway, Desmarais, Harden, LaCombe,
  Linder, and Wallach]{boehmke2019spid_data}
Boehmke, F., Brockway, M., Desmarais, B., Harden, J.~J., LaCombe, S., Linder,
  F., and Wallach, H.
\newblock {State Diffusion Networks - Latent Network Ties from SPID v1.0},
  2019.
\newblock URL \url{https://doi.org/10.7910/DVN/1QJCDJ}.

\bibitem[Boehmke et~al.(2018)Boehmke, Brockway, Desmarais, Harden, LaCombe,
  Linder, and Wallach]{boehmke2018new}
Boehmke, F.~J., Brockway, M., Desmarais, B.~A., Harden, J.~J., LaCombe, S.,
  Linder, F., and Wallach, H.
\newblock A new database for inferring public policy innovativeness and
  diffusion networks.
\newblock \emph{Available at SSRN 3199383}, 2018.

\bibitem[Bollob{\'a}s et~al.(2003)Bollob{\'a}s, Borgs, Chayes, and
  Riordan]{bollobas2003directed}
Bollob{\'a}s, B., Borgs, C., Chayes, J., and Riordan, O.
\newblock Directed scale-free graphs.
\newblock In \emph{Proceedings of the fourteenth annual ACM-SIAM symposium on
  Discrete algorithms}, pp.\  132--139. Society for Industrial and Applied
  Mathematics, 2003.

\bibitem[Byrd et~al.(1995)Byrd, Lu, Nocedal, and Zhu]{byrd1995limited}
Byrd, R.~H., Lu, P., Nocedal, J., and Zhu, C.
\newblock A limited memory algorithm for bound constrained optimization.
\newblock \emph{SIAM Journal on scientific computing}, 16\penalty0
  (5):\penalty0 1190--1208, 1995.

\bibitem[Chen et~al.(2018)Chen, Rubanova, Bettencourt, and
  Duvenaud]{chen2018neural}
Chen, T.~Q., Rubanova, Y., Bettencourt, J., and Duvenaud, D.~K.
\newblock Neural ordinary differential equations.
\newblock In \emph{Advances in neural information processing systems}, pp.\
  6571--6583, 2018.

\bibitem[Daley \& Vere-Jones(2008)Daley and Vere-Jones]{daley2008introduction}
Daley, D.~J. and Vere-Jones, D.
\newblock An introduction to the theory of point processes.
\newblock \emph{Probability and Its Applications}, 2008.
\newblock ISSN 1431-7028.
\newblock \doi{10.1007/978-0-387-49835-5}.

\bibitem[Du et~al.(2013)Du, Song, Gomez{-}Rodriguez, and Zha]{du2013scalable}
Du, N., Song, L., Gomez{-}Rodriguez, M., and Zha, H.
\newblock Scalable influence estimation in continuous-time diffusion networks.
\newblock In \emph{Advances in Neural Information Processing Systems 26: 27th
  Annual Conference on Neural Information Processing Systems 2013. Proceedings
  of a meeting held December 5-8, 2013, Lake Tahoe, Nevada, United States},
  pp.\  3147--3155, 2013.

\bibitem[Du et~al.(2016)Du, Dai, Trivedi, Upadhyay, Gomez-Rodriguez, and
  Song]{du2016recurrent}
Du, N., Dai, H., Trivedi, R., Upadhyay, U., Gomez-Rodriguez, M., and Song, L.
\newblock Recurrent marked temporal point processes: Embedding event history to
  vector.
\newblock In \emph{Proceedings of the 22nd ACM SIGKDD International Conference
  on Knowledge Discovery and Data Mining}, pp.\  1555--1564, 2016.

\bibitem[Farajtabar et~al.(2017{\natexlab{a}})Farajtabar, Wang,
  Gomez-Rodriguez, Li, Zha, and Song]{farajtabar2017coevolve}
Farajtabar, M., Wang, Y., Gomez-Rodriguez, M., Li, S., Zha, H., and Song, L.
\newblock Coevolve: A joint point process model for information diffusion and
  network evolution.
\newblock \emph{The Journal of Machine Learning Research}, 18\penalty0
  (1):\penalty0 1305--1353, 2017{\natexlab{a}}.

\bibitem[Farajtabar et~al.(2017{\natexlab{b}})Farajtabar, Yang, Ye, Xu,
  Trivedi, Khalil, Li, Song, and Zha]{farajtabar2017fakenews}
Farajtabar, M., Yang, J., Ye, X., Xu, H., Trivedi, R., Khalil, E.~B., Li, S.,
  Song, L., and Zha, H.
\newblock Fake news mitigation via point process based intervention.
\newblock In \emph{Proceedings of the 34th International Conference on Machine
  Learning, {ICML} 2017, Sydney, NSW, Australia, 6-11 August 2017}, pp.\
  1097--1106, 2017{\natexlab{b}}.

\bibitem[Gomez{-}Rodriguez et~al.(2011)Gomez{-}Rodriguez, Balduzzi, and
  Sch{\"{o}}lkopf]{gomez-rodriguez2011uncovering}
Gomez{-}Rodriguez, M., Balduzzi, D., and Sch{\"{o}}lkopf, B.
\newblock Uncovering the temporal dynamics of diffusion networks.
\newblock In \emph{Proceedings of the 28th International Conference on Machine
  Learning, {ICML} 2011, Bellevue, Washington, USA, June 28 - July 2, 2011},
  pp.\  561--568, 2011.

\bibitem[Gomez~Rodriguez et~al.(2013)Gomez~Rodriguez, Leskovec, and
  Schölkopf]{gomez-rodriguez2013structure}
Gomez~Rodriguez, M., Leskovec, J., and Schölkopf, B.
\newblock Structure and dynamics of information pathways in online media.
\newblock \emph{Proceedings of the sixth ACM international conference on Web
  search and data mining - WSDM ’13}, 2013.
\newblock \doi{10.1145/2433396.2433402}.
\newblock URL \url{http://snap.stanford.edu/infopath/data.html}.

\bibitem[Hawkes(1971)]{hawkes1971spectra}
Hawkes, A.~G.
\newblock Spectra of some self-exciting and mutually exciting point processes.
\newblock \emph{Biometrika}, 58\penalty0 (1):\penalty0 83–90, 1971.
\newblock ISSN 1464-3510.
\newblock \doi{10.1093/biomet/58.1.83}.

\bibitem[He et~al.(2015)He, Rekatsinas, Foulds, Getoor, and
  Liu]{he2015hawkestopic}
He, X., Rekatsinas, T., Foulds, J., Getoor, L., and Liu, Y.
\newblock Hawkestopic: A joint model for network inference and topic modeling
  from text-based cascades.
\newblock In \emph{International conference on machine learning}, pp.\
  871--880, 2015.

\bibitem[Iwata et~al.(2013)Iwata, Shah, and Ghahramani]{iwata2013discovering}
Iwata, T., Shah, A., and Ghahramani, Z.
\newblock Discovering latent influence in online social activities via shared
  cascade poisson processes.
\newblock In \emph{Proceedings of the 19th ACM SIGKDD international conference
  on Knowledge discovery and data mining}, pp.\  266--274, 2013.

\bibitem[Jankowiak \& Gomez-Rodriguez(2017)Jankowiak and
  Gomez-Rodriguez]{jankowiak2017uncovering}
Jankowiak, M. and Gomez-Rodriguez, M.
\newblock Uncovering the spatiotemporal patterns of collective social activity.
\newblock In \emph{Proceedings of the 2017 SIAM International Conference on
  Data Mining}, pp.\  822--830. SIAM, 2017.

\bibitem[Jia \& Benson(2019)Jia and Benson]{jia2019neuraljump}
Jia, J. and Benson, A.~R.
\newblock Neural jump stochastic differential equations.
\newblock In \emph{Advances in Neural Information Processing Systems 32: Annual
  Conference on Neural Information Processing Systems 2019, NeurIPS 2019, 8-14
  December 2019, Vancouver, BC, Canada}, pp.\  9843--9854, 2019.

\bibitem[Jing \& Smola(2017)Jing and Smola]{jing2017survival}
Jing, H. and Smola, A.~J.
\newblock Neural survival recommender.
\newblock In \emph{Proceedings of the Tenth ACM International Conference on Web
  Search and Data Mining}, WSDM ’17, pp.\  515–524, New York, NY, USA,
  2017. Association for Computing Machinery.
\newblock ISBN 9781450346757.
\newblock \doi{10.1145/3018661.3018719}.

\bibitem[Kingma \& Ba(2015)Kingma and Ba]{kingma2014adam}
Kingma, D.~P. and Ba, J.
\newblock Adam: {A} method for stochastic optimization.
\newblock In \emph{3rd International Conference on Learning Representations,
  {ICLR} 2015, San Diego, CA, USA, May 7-9, 2015, Conference Track
  Proceedings}, 2015.

\bibitem[Kumar et~al.(2018)Kumar, Hamilton, Leskovec, and
  Jurafsky]{kumar2018community}
Kumar, S., Hamilton, W.~L., Leskovec, J., and Jurafsky, D.
\newblock Community interaction and conflict on the web.
\newblock In \emph{Proceedings of the 2018 World Wide Web Conference on World
  Wide Web}, pp.\  933--943. International World Wide Web Conferences Steering
  Committee, 2018.
\newblock URL \url{https://snap.stanford.edu/data/soc-RedditHyperlinks.html}.

\bibitem[Kumar et~al.(2019)Kumar, Zhang, and Leskovec]{kumar2019predicting}
Kumar, S., Zhang, X., and Leskovec, J.
\newblock Predicting dynamic embedding trajectory in temporal interaction
  networks.
\newblock In \emph{Proceedings of the 25th ACM SIGKDD international conference
  on Knowledge discovery and data mining}. ACM, 2019.

\bibitem[Lemonnier et~al.(2017)Lemonnier, Scaman, and
  Kalogeratos]{lemonnier2017multivariate}
Lemonnier, R., Scaman, K., and Kalogeratos, A.
\newblock Multivariate {Hawkes} processes for large-scale inference.
\newblock In \emph{Proceedings of the Thirty-First {AAAI} Conference on
  Artificial Intelligence, February 4-9, 2017, San Francisco, California,
  {USA}}, pp.\  2168--2174, 2017.

\bibitem[Leskovec et~al.(2009)Leskovec, Backstrom, and
  Kleinberg]{leskovec2009meme}
Leskovec, J., Backstrom, L., and Kleinberg, J.
\newblock Meme-tracking and the dynamics of the news cycle.
\newblock In \emph{Proceedings of the 15th ACM SIGKDD international conference
  on Knowledge discovery and data mining}, pp.\  497--506, 2009.
\newblock URL \url{https://snap.stanford.edu/data/memetracker9.html}.

\bibitem[Linderman \& Adams(2014)Linderman and Adams]{linderman2014discovering}
Linderman, S. and Adams, R.
\newblock Discovering latent network structure in point process data.
\newblock In \emph{International Conference on Machine Learning}, pp.\
  1413--1421, 2014.

\bibitem[Mavroforakis et~al.(2017)Mavroforakis, Valera, and
  Gomez-Rodriguez]{mavroforakis2017modeling}
Mavroforakis, C., Valera, I., and Gomez-Rodriguez, M.
\newblock Modeling the dynamics of learning activity on the web.
\newblock In \emph{Proceedings of the 26th International Conference on World
  Wide Web}, pp.\  1421--1430, 2017.

\bibitem[Mei \& Eisner(2017)Mei and Eisner]{tpp/mei2017neuralhawkes}
Mei, H. and Eisner, J.
\newblock The {Neural Hawkes Process}: {A} neurally self-modulating
  multivariate point process.
\newblock In \emph{Advances in Neural Information Processing Systems 30: Annual
  Conference on Neural Information Processing Systems 2017, 4-9 December 2017,
  Long Beach, CA, {USA}}, pp.\  6754--6764, 2017.

\bibitem[Ogata(1981)]{ogata1981lewis}
Ogata, Y.
\newblock On lewis' simulation method for point processes.
\newblock \emph{IEEE transactions on information theory}, 27\penalty0
  (1):\penalty0 23--31, 1981.

\bibitem[Ogata(1998)]{ogata1998space}
Ogata, Y.
\newblock Space-time point-process models for earthquake occurrences.
\newblock \emph{Annals of the Institute of Statistical Mathematics},
  50\penalty0 (2):\penalty0 379--402, 1998.

\bibitem[Ozaki(1979)]{tpp/ozaki1979maximum}
Ozaki, T.
\newblock Maximum likelihood estimation of {Hawkes’} self-exciting point
  processes.
\newblock \emph{Annals of the Institute of Statistical Mathematics},
  31\penalty0 (1):\penalty0 145–155, Dec 1979.
\newblock ISSN 1572-9052.
\newblock \doi{10.1007/bf02480272}.

\bibitem[Recht et~al.(2011)Recht, R{\'{e}}, Wright, and
  Niu]{DBLP:conf/nips/RechtRWN11}
Recht, B., R{\'{e}}, C., Wright, S.~J., and Niu, F.
\newblock Hogwild: {A} lock-free approach to parallelizing stochastic gradient
  descent.
\newblock In \emph{Advances in Neural Information Processing Systems 24}, pp.\
  693--701, 2011.

\bibitem[Rizoiu et~al.(2018)Rizoiu, Mishra, Kong, Carman, and
  Xie]{rizoiu2018sir}
Rizoiu, M.-A., Mishra, S., Kong, Q., Carman, M., and Xie, L.
\newblock Sir-hawkes: linking epidemic models and hawkes processes to model
  diffusions in finite populations.
\newblock In \emph{Proceedings of the 2018 World Wide Web Conference}, pp.\
  419--428, 2018.

\bibitem[Rubanova et~al.(2019)Rubanova, Chen, and
  Duvenaud]{rubanova2019latentodes}
Rubanova, Y., Chen, R. T.~Q., and Duvenaud, D.
\newblock Latent {ODEs} for irregularly-sampled time series.
\newblock \emph{CoRR}, abs/1907.03907, 2019.

\bibitem[Shchur et~al.(2019)Shchur, Bilos, and
  G{\"{u}}nnemann]{shchur2019intensityfree}
Shchur, O., Bilos, M., and G{\"{u}}nnemann, S.
\newblock Intensity-free learning of temporal point processes.
\newblock \emph{CoRR}, abs/1909.12127, 2019.

\bibitem[Xiao et~al.(2017)Xiao, Farajtabar, Ye, Yan, Yang, Song, and
  Zha]{xiao2017wassersteinpp}
Xiao, S., Farajtabar, M., Ye, X., Yan, J., Yang, X., Song, L., and Zha, H.
\newblock Wasserstein learning of deep generative point process models.
\newblock In \emph{Advances in Neural Information Processing Systems 30: Annual
  Conference on Neural Information Processing Systems 2017, 4-9 December 2017,
  Long Beach, CA, {USA}}, pp.\  3247--3257, 2017.

\bibitem[Xu et~al.(2016)Xu, Farajtabar, and Zha]{xu2016grangermhp}
Xu, H., Farajtabar, M., and Zha, H.
\newblock Learning {Granger} causality for {Hawkes} processes.
\newblock In \emph{Proceedings of the 33nd International Conference on Machine
  Learning, {ICML} 2016, New York City, NY, USA, June 19-24, 2016}, pp.\
  1717--1726, 2016.

\bibitem[Zhao et~al.(2015)Zhao, Erdogdu, He, Rajaraman, and
  Leskovec]{zhao2015seismic}
Zhao, Q., Erdogdu, M.~A., He, H.~Y., Rajaraman, A., and Leskovec, J.
\newblock Seismic: A self-exciting point process model for predicting tweet
  popularity.
\newblock In \emph{Proceedings of the 21th ACM SIGKDD International Conference
  on Knowledge Discovery and Data Mining}, pp.\  1513--1522, 2015.

\bibitem[Zhou et~al.(2013)Zhou, Zha, and Song]{zhou2013learning}
Zhou, K., Zha, H., and Song, L.
\newblock Learning social infectivity in sparse low-rank networks using
  multi-dimensional {Hawkes} processes.
\newblock In \emph{Artificial Intelligence and Statistics}, pp.\  641--649,
  2013.

\end{thebibliography}
\end{document}